\documentclass{article}

\usepackage{arxiv}

\usepackage[utf8]{inputenc} % allow utf-8 input
\usepackage[T1]{fontenc}% use 8-bit T1 fonts
\usepackage{hyperref} % hyperlinks
\usepackage{url}% simple URL typesetting
\usepackage{booktabs} % professional-quality tables
\usepackage{amsfonts} % blackboard math symbols
\usepackage{nicefrac} % compact symbols for 1/2, etc.
\usepackage{microtype}% microtypography
\usepackage[title]{appendix}
\usepackage{graphicx}
\usepackage[bf]{caption}
\usepackage{diagbox}
\usepackage{float}% table position
\usepackage{multirow}% for \multirow command
\usepackage{tabularx}

\usepackage{doi}
\usepackage{etoolbox}
\usepackage[square,numbers]{natbib}

\newtoggle{bibdoi}
\newtoggle{biburl}
\makeatletter
\newsavebox{\bib@url}
\newsavebox{\bib@doi}

\undef{\APACrefURL}
\undef{\endAPACrefURL}
\undef{\APACrefDOI}
\undef{\endAPACrefDOI}

\newenvironment{APACrefURL}
{\global\toggletrue{biburl}\lrbox\bib@url}
{\endlrbox}

\newenvironment{APACrefDOI}
{\global\toggletrue{bibdoi}\lrbox\bib@doi}
{\endlrbox}

\newcommand{\printinfo}{
\iftoggle{bibdoi}{\usebox{\bib@doi}}{\usebox{\bib@url}}
\togglefalse{bibdoi}
}

\AtBeginEnvironment{thebibliography}{
\pretocmd{\PrintBackRefs}{%
\iftoggle{bibdoi}
{\iftoggle{biburl}{\unskip\unskip}{}\usebox{\bib@doi}}
{\iftoggle{biburl}{Retrieved from \usebox{\bib@url}}}{}
\togglefalse{bibdoi}\togglefalse{biburl}%
}{}{}}
\makeatother
\title{ML4EJ: Decoding the Role of Urban Features in Shaping Environmental Injustice Using Interpretable Machine Learning}
\date{} 					% Or removing it

\renewcommand{\headeright}{}
\renewcommand{\undertitle}{}
\renewcommand{\shorttitle}{}

\hypersetup{
pdftitle={ML4EJ: Decoding the Role of Urban Features in Shaping Environmental Injustice Using Interpretable Machine Learning},
pdfsubject={},
}

\begin{document}
\maketitle

\begin{center}
% authors go here:
{\Large
Yu-Hsuan Ho\textsuperscript{a},
Zhewei Liu\textsuperscript{a,*},
Cheng-Chun Lee\textsuperscript{a}, 
Ali Mostafavi\textsuperscript{a}
\par}

\bigskip
\textsuperscript{a} Urban Resilience.AI Lab, Zachry Department of Civil and Environmental Engineering,\\ Texas A\&M University, College Station, TX\\
\vspace{6pt}
\textsuperscript{*} correseponding author, email: zheweiliu@tamu.edu
\\
\end{center}
\bigskip
\begin{abstract}

Understanding the key factors shaping environmental hazard exposures and their associated environmental injustice issues is vital for formulating equitable policy measures. Traditional perspectives on environmental injustice have primarily focused on the socioeconomic dimensions, often overlooking the influence of heterogeneous urban characteristics. This limited view may obstruct a comprehensive understanding of the complex nature of environmental justice and its relationship with urban design features. To address this gap, this study creates an interpretable machine learning model to examine the effects of various urban features and their non-linear interactions to the exposure disparities of three primary hazards: air pollution, urban heat, and flooding. The analysis trains and tests models with data from six metropolitan counties in the United States using Random Forest and XGBoost. The predictability performance of the models is used to measure the extent to which variations of urban features shape disparities in environmental hazard levels. In addition, the analysis of feature importance reveals features related to social-demographic characteristics as the most prominent urban features that shape environmental hazard extent. Features related to transportation infrastructures and facility distribution as the relatively important urban features for urban heat exposure, and features related to land cover as the relatively important urban features for air pollution exposure. The importance of these features varies between regions. In addition, we evaluate the models' transferability across different regions and hazards. The results highlight limited transferability, underscoring the intricate differences among hazards and regions and the way in which urban features shape environmental hazard exposures. The insights gleaned from this study offer fresh perspectives on the relationship among urban features and their interplay with environmental hazard exposure disparities, informing the development of more integrated urban design policies to enhance social equity and environmental injustice issues.

\end{abstract}

% keywords can be removed
\keywords{environmental injustice \and urban inequality \and integrated urban design \and machine learning \and equity}

% =======note=======
% \citep{}: with ()
% \citet{}: author name only
% \ref{fig:fig1}: figure
% \begin{figure}
% 	\centering
% \includegraphics[width=0.85\linewidth]{fig1.png}
% \caption{Schematic .}
% 	\label{fig:fig1}
% \end{figure}
% \ref{sec:Results}: section
% \section{Results}
% \label{sec:Results}
% section hierarchy
% \section{}
% \subsection{}
% \subsubsection{}
% =======note=======

\section{Introduction}
\label{sec:1}
Environmental justice issues embody the unequal distribution of environmental hazards and benefits, often along lines of race and socio-economic status. Specifically, x`vulnerable communities (e.g., minority and economically disadvantageous) bear significant disproportionate overexposure to the various environmental hazards \cite{hallegatte2016unbreakable,liu2023collision}. These issues raise critical questions about equity, public health, and sustainable development, necessitating further investigation into the factors shaping environmental injustice in communities. Comprehending the determinants of environmental justice issues is not only a matter of academic interest but a moral and policy imperative that can guide the development of effective policies and strategies to alleviate these injustices.

The existing body of research has made significant strides towards unpacking the determinants of environmental injustice, shedding light on the economic, sociopolitical, and racial factors that underpin the unequal distribution of environmental hazards. However, despite these valuable contributions, a conspicuous gap persists in our understanding of the role and impact of specific urban features in shaping these inequities \cite{watson2015beyond,calderon2021tracing}. While it is widely recognized that urban planning and design can greatly influence the living experiences of residents, remarkably few studies have delved into the direct relationship between such urban features and the extent of environmental hazard exposure and its disparity \cite{soja2013seeking,mendez2020visible}. This lack of focus could obscure critical dynamics that feed into systemic environmental injustice. In particular, understanding how urban features such as built environment and land cover contribute to the environmental injustice outcomes may hold the key to integrated urban design strategies that promote environmental justice, health, and sustainability \cite{Pulido2000environmental,Schell2020ecological,Flocks2011treecover,YIN2023103687,liu2023residence}. Moreover, deciphering the commonality of cross-region environmental injustice sheds light on the homogeneity in broader geographical scope, and provides generalizable insights for equitable decision making. This gap in the literature presents an opportunity for further research that explicitly investigates how urban features contribute to environmental hazard exposure or could be used to mitigate environmental injustice.

To untangle the relationship between diverse urban features and environmental hazard levels, we employ interpretable machine learning techniques. In this study, six counties in the United States with similar population size are selected as study regions. To decode the relationship between environmental hazards and heterogeneous urban features, we employ interpretable machine learning models to handle high-dimensional data, capturing complex and non-linear relationships among urban features. In addition, the analysis investigates the transferability of the trained models across hazards types and across cities, examining features that commonly shape environmental injustice issues across different cities and hazard types, to inform integrated urban design policies. Using the results of the machine learning models, this study aims to answer the following interrelated research questions:

\begin{itemize}
\item RQ1: To what extent do various urban features influence the spatial exposure levels to environmental hazards? What urban features play a more prominent role in shaping environmental hazard extent?
\item RQ2: What are the common and distinct urban features that contribute to spatial exposure levels to a variety of environmental hazards?
\item RQ3: To what extent do the determinants of environmental hazards vary across different regions? How transferable are policies addressing these determinants from one region to another?
\end{itemize}

\begin{figure}[ht]
\centering
\includegraphics[width=1\textwidth]{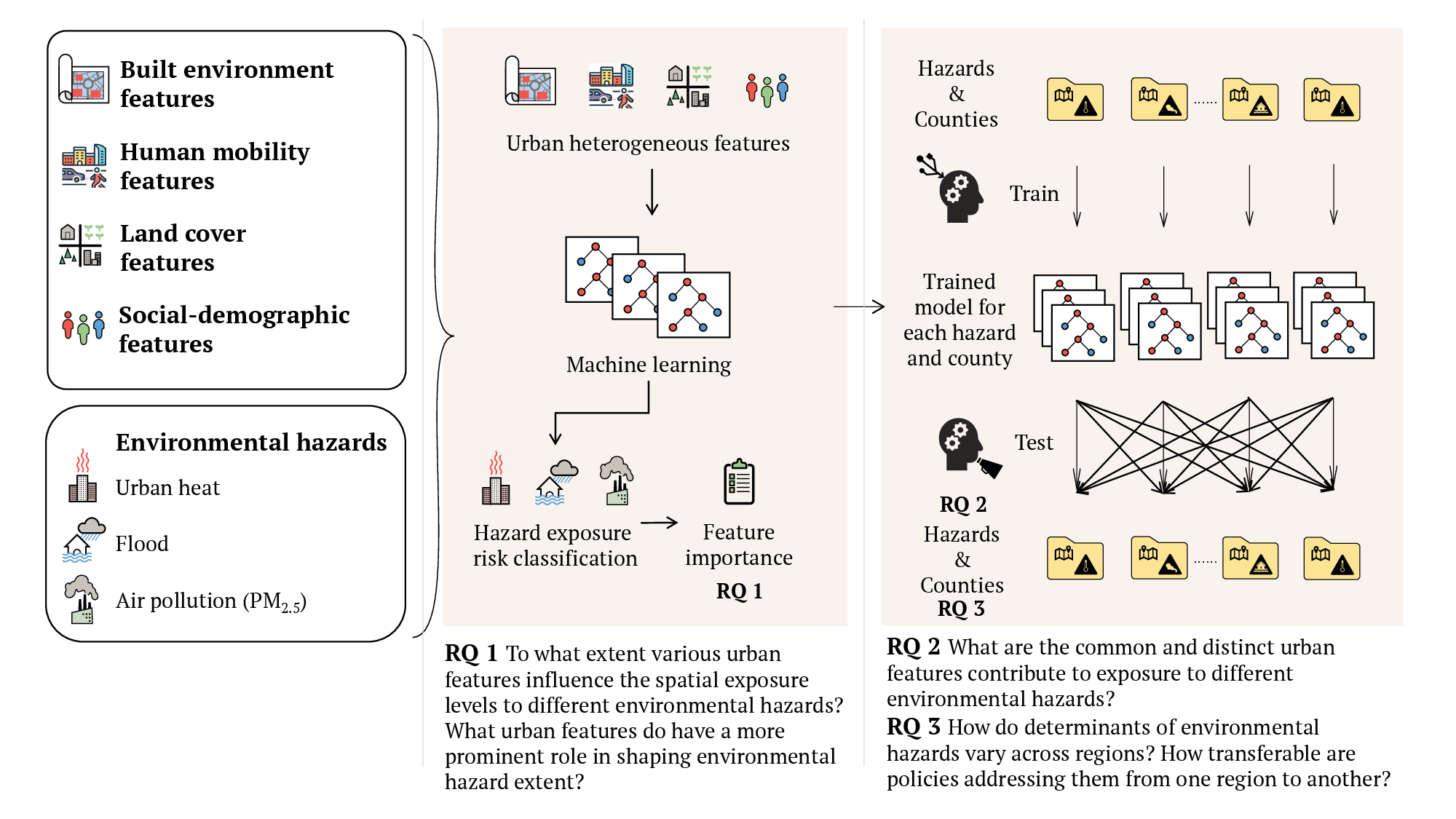}
\caption{Study overview: The urban features constructed from datasets related to social-demographics, built environment, human mobility, and land cover are used as input features. Interpretable machine learning models are adopted for predicting the extent of spatial hazards exposure. Feature importance analysis is based on interpretations of the trained machine learning models (RQ1). By applying the model trained for one hazard or county to another, the transferability of models is evaluated to answer RQ 2 and RQ 3.}
\label{fig:conceptual_figure}
\end{figure}

Figure \ref{fig:conceptual_figure} depicts an overview of this study. First, representations of the various urban features are constructed from a wide range of data sources, including social-demographics, built environment, human mobility, and land cover. Then interpretable machine learning models are created to predict spatial exposure levels to three primary environmental hazards: air pollution, urban heat, and flooding. The model's predictability performance serves as a measurement of environmental injustice, since it suggests the extent to which hazards exposure is shaped by urban features. Then the important features for each hazard type based on interpretations of the trained machine learning models are revealed. Subsequently, we test the model's transferability by applying the model trained for one city or hazard to another city or hazard. Our study not only quantifies the environmental injustice based on model predictability, but also pinpoints the urban features contributing to the such inequalities. Furthermore, the model outputs examine the similarity and variations of these inequalities across different regions and hazards, thereby providing evidence of the extent to which policies for alleviating environmental injustice could be transferable across hazard types and regions. 

The findings of the study offer multiple important contributions to various academic disciplines and diverse stakeholder practitioners: (1) the ability of machine learning models to predict environmental hazard extent provides urban scientists and planners with tools to capture the complex and non-linear interactions among urban features that shape disparities in spatial exposure to environmental hazards; (2) the findings reveal common and distinct urban features that shape spatial exposure to different hazard types; this finding informs urban planners and environmental managers regarding effective integrated urban design strategies that could alleviate multiple environmental hazards and inequalities;(3) The results highlight the importance of examining intertwined urban features in evaluating complex and emergent urban phenomena; accordingly, the interpretable machine learning approach adopted in this study could inform future studies in urban computing and complexity fields in studying urban phenomena such as environmental justice issues, urban health, and sustainability. 

\section{Data and Methodology}
\label{sec:2}
\subsection{Study Area and Data Description}
For the study areas, we selected six U.S. metropolitan counties in different regions based on their similarity and population size: Harris County in Texas, Cook County in Illinois, Wayne County in Michigan,Suffolk County in Massachusetts, Fulton County in Georgia, and Queens County in New York. For all study areas, the study examined urban features constructed from various datasets.

Table \ref{Tab:Feature} presents the constructed features investigated in this study. The features are derived from urban characteristics, including built environment, human mobility, land cover, and social-demographics, which may have profound impact on the extent of environmental hazards. The built environment features capture the human-made building surroundings, including facilities, transportation infrastructure, and public spaces, which can influence people’s behavior and their exposure to environmental hazards\cite{GRINESKI201225,WOLCH2014234}. Human mobility features capture people’s movement from one place to another, revealing travel patterns that can be affected by reduced accessibility to facilities and exposure to environmental hazards \cite{fondelli2008fine}. Land cover features capture the physical and biological characteristics of the land surface, which can have significant effects on the impacts of environmental hazards \cite{LIN2023104464,brody2008identifying}. Social-demographics refers to the characteristics of a population, such as income, race, age, and education, which is crucial to explore whether vulnerable populations are exposed to higher level of environmental hazards \cite{Otto2017SocialVT}. The environmental hazards investigated in this work are urban heat, flood risk, and air pollution (PM\(_{2.5}\)), presented in Table \ref{Tab:Hazard}. 

\begin{figure}[ht]
\centering
\includegraphics[width=1\textwidth]{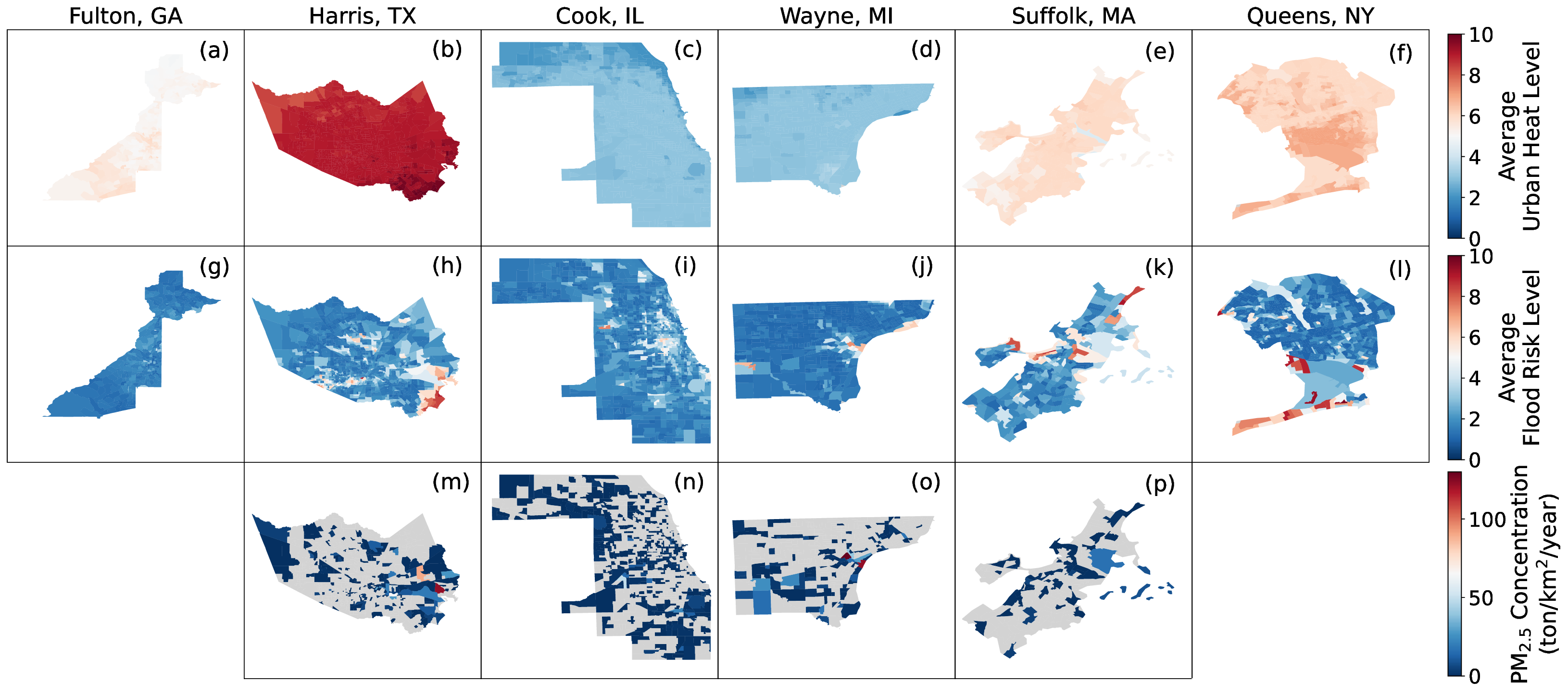}
\caption{Spatial distribution of the exposure of environmental hazards. (a)-(f) Distributions of urban heat risk level in Fulton County, Harris County, Cook County, Wayne County, Suffolk County, and Queens County. (g)-(l) Distributions of flood risk levels in respective counties. (m)-(p) Distributions of the concentration of PM\(_{2.5}\). The urban heat risk levels and flood risk levels are the average value for all households within each census tract.}
\label{fig:hazards_map}
\end{figure}

Figure \ref{fig:hazards_map} shows the spatial distribution of the investigated environmental hazards in the study regions. We analyzed environmental hazard exposure levels at the spatial unit level of the census tract. The spatial patterns show great variation in environmental hazard exposures, particularly urban heat, among the counties. Since the units for spatial extent of each environmental hazard type vary, we classified model output into high-risk or low-risk areas. The areas with hazard exposure greater than the mean value is classified as high-risk, and the rest as low-risk.

\begin{table}
\caption{Feature Description}
\centering
\begin{tabularx}{\textwidth} {
>{\hsize=.27\hsize\raggedright\arraybackslash}X
>{\hsize=.65\hsize\raggedright\arraybackslash}X
>{\hsize=.10\hsize\raggedright\arraybackslash}X
>{\hsize=.08\hsize\raggedright\arraybackslash}X
}
\cline{1-4}
Feature & \multicolumn{1}{c}{Description}& \multicolumn{1}{c}{Source} & \multicolumn{1}{c}{Literature}\\
\cline{1-4}
\multicolumn{4}{l}{\textbf{Built environment features}}\\
POI\_Density& Number of all points of interest (POIs) per square kilometers& \multirow{5}{\hsize}{SafeGraph} & \multirow{5}{\hsize}{\cite{GRINESKI201225,liu2022graph,wing2000hog}}\\
Grocery\_Density& Number of grocery stores per square kilometers && \\
Health\_Care\_Density & Number of health care facilities per square kilometers && \\
Recreation\_Gym\_Density& Number of recreation and gym centers per square kilometers && \\
Restaurant\_Density& Number of restaurants per square kilometers && \\
\cline{3-4}
Walkability &National Walkability Index from U.S. Environmental Protection Agency & \multirow{7}{\hsize}{Agency for Toxic Substances and Disease Registry (ATSDR)}& \multirow{4}{\hsize}{\cite{YIN2023103687,cutts2009city,JEON2023104631}} \\
High\_Volume\_Road& Proportion of area within 1-mi buffer of a high-volume street or road&& \\
Railway & Proportion of area within 1-mi buffer of railways && \\
Airport & Proportion of area within 1-mi buffer of airports&& \\
\cline{4-4}
Park\_Lack& Proportion of area not within 1-mi buffer of parks, recreational areas, or public forests && \multirow{3}{\hsize}{\cite{WOLCH2014234,wing2000hog,CHIANG2023104679,wilson2008water}}\\
Impaired\_Water & Percent of watershed area classified as impaired && \\
Building\_Age & Proportion of occupied housing units built prior to 1980 && \\
\cline{1-4}
\multicolumn{4}{l}{\textbf{Human mobility features}}\\
POI\_Avg\_Time & Average travel time (min) at individual level to all POIs per day & \multirow{4}{\hsize}{Spectus} & \multirow{4}{\hsize}{\cite{liu2023residence,liu2022graph}}\\
POI\_Avg\_Distance & Average travel distance (m) at individual level to all POIs per day && \\
POI\_Avg\_Trips& Average number of trips at individual level to all POIs per day && \\
POI\_Avg\_ROG& Average travel radius of gyration (ROG) (m) at individual level to all POIs per day && \\
\cline{1-4}
\multicolumn{4}{l}{\textbf{Land cover features}}\\
Low\_Developed& Proportion of areas developed in low intensity (urban imperviousness 20-49\%)& \multirow{3}{\hsize}{National Neighborhood Data Archive (NaNDA)}& \multirow{3}{\hsize}{\cite{LIN2023104464,GONG2020111510}} \\
Medium\_Developed & Proportion of areas developed in medium intensity (urban imperviousness 50-79\%) && \\
High\_Developed & Proportion of areas developed in high intensity (urban imperviousness \textgreater 79\%) && \\
\cline{3-4}
Average\_Tree\_Canopy & Average percentage of areas covered by tree canopy, estimates in each census tract & \multirow{2}{\hsize}{Multi-Resolution Land Characteristics (MRLC)}& \\
Maximum\_Tree\_Canopy & Maximum percentage of areas covered by tree canopy, estimates in each census tract &&\multirow{1}{\hsize}{\cite{Flocks2011treecover,LIN2023104464,ZHOU20171}} \\\\\\
\cline{1-4}
\multicolumn{4}{l}{\textbf{Social-demographic features}}\\
Income& Mean household income& \multirow{3}{\hsize}{U.S. Census Bureau}& \multirow{16}{\hsize}{\cite{hallegatte2016unbreakable,liu2023collision,anderton1994environmental,pastor2001came,pais2014unequal,morello2006separate,mendez2020visible,Pulido2000environmental,Schell2020ecological,GRINESKI201225,WOLCH2014234,liu2022graph,wing2000hog}} \\
Race& Percentage of non-white population && \\
Population\_Density & Total population per square kilometers && \\
\cline{3-3}
Poverty & Percentage of population with income below 200\% of federal poverty level& \multirow{13}{\hsize}{Agency for Toxic Substances and Disease Registry (ATSDR)} & \\
Education & Percentage of population aged 25 and older with no high school diploma && \\
Unemployment& percentage of population age 16 and older who are unemployed && \\
Renter\_Occupied\_Housing & Percentage of housing units that are renter-occupied && \\
Housing\_Burden & Percentage of households burdened by housing costs (pay greater than 30\% of monthly income on housing expenses) && \\
Group\_Quarter& Percentage of persons living in group quarters && \\
Mobile\_Home& Percentage of total housing units designated as mobile homes && \\
Health\_Insurance\_Lack & Percentage of population without health insurance&& \\
Internet\_Lack& Percentage of households without internet subscription && \\
Elder &percentage of population aged 65 and older && \\
Children& Percentage of population aged 17 and younger && \\
Disability& Percentage of civilian, noninstitutionalized population with a disability&& \\
Limited\_English\_Speaking& Percentage of population aged 5 and with lesser spoken English proficiency && \\ 
\cline{1-4}
\end{tabularx}
\label{Tab:Feature}
\end{table}

\begin{table}
\caption{Hazard Description}
\centering

\begin{tabularx}{\textwidth} {
>{\hsize=.15\hsize\raggedright\arraybackslash}X
>{\hsize=.65\hsize\raggedright\arraybackslash}X
>{\hsize=.15\hsize\raggedright\arraybackslash}X
>{\hsize=.1\hsize\raggedright\arraybackslash}X
}
\cline{1-4}
Hazard & \multicolumn{1}{c}{Description} & \multicolumn{1}{c}{Source} & \multicolumn{1}{c}{Literature} \\
\cline{1-4}\\
Air pollution & Air pollution refers to the particles and gases in the air with negative impacts on human health and the environment. In this study, the concentration of PM\(_{2.5}\), one of the criteria pollutants to evaluate air quality, is selected to represent air pollution. & Environmental Protection Agency (EPA) & \cite{liu2023residence,liu2022graph,jerrett2001gis,anderson2018climate,wang2023network} \\\\
Urban heat & heat risk for a property is expressed as a numerical rating from 1 (minimal risk) to 10 (extreme risk), representing the property's risk of exposure to extreme heat in the coming three decades, which is calculated by First Street Foundation Extreme Heat Model (FSF-EHM), considering the current average of daily maximum temperature and humidity at the location of the property, as well as the projected increase in the average of daily maximum temperature and humidity for that location over the next three decades \cite{firststreet2023heat}. & First Street Foundation & \cite{liu2023residence,GRINESKI201225,LIN2023104464,ZHOU20171} \\\\
Flood & Flood risk for a property is expressed as a numerical rating from 1 (minimal risk) to 10 (extreme risk) representing the property's cumulative risk of experiencing flooding throughout a 30-year mortgage period, which is calculated by the First Street Foundation Flood Model (FSF-FM), taking into account expected environmental changes, such as rising sea levels, evolving precipitation patterns, and increasing temperatures in both sea surfaces and the atmosphere \cite{firststreet2020flood}. & First Street Foundation & \cite{GRINESKI201225,maantay2009mapping,chakraborty2014social,ho2023elevvision} \\

\cline{1-4}
\end{tabularx}
\label{Tab:Hazard}
\end{table}

\subsection{Model Training}
Machine learning is recognized for its ability to effectively capture the complex interactions among input features that influence the output variable \citep{fan_inequality_2023,liu2019recommending,shi2021regnet,liu2023decoding}. Tree-based models are particularly advantageous over other machine learning models in terms of interpretability and robustness against uninformative features \cite{Grinsztajn2022treebased}. In this study, random forest and XGBoost, two effective interpretable tree-based machine learning models, are adopted to explore the nonlinear relationships among heterogeneous urban features in predicting the spatial extent of environmental hazards exposure. The predictability performance of models measures the extent of the relationship between variations of urban features and disparities in environmental hazard levels. Interpretation of the output of these models offer insights into the contribution of urban features to environmental hazard exposure levels \cite{hough2004assessing}. The dataset is divided into a training set (70\%) and a test set (30\%). In training process, ten-fold cross-validation is used for hyperparameter tuning.

\subsection{Feature Importance Analysis}
% https://www.jianshu.com/p/7a876bb876b9
In this study, we use normalized Gini importance to analyze feature importance. This metric offers two main advantages: (1) computational efficiency and (2) straightforward comparisons between feature importance due to its standardized scales. The formula of normalized Gini importance is defined as Eq. \ref{eq:gini},

\begin{equation}\label{eq:gini}
G = \sum_{k=1}^{K} p_k(1-p_k)
\end{equation}

where \(K\) is the number of classes and \(p_k\) is the ratio of class \(k\). The importance of node \(m\) is calculated as Eq. \ref{eq:imp},

\begin{equation}\label{eq:imp}
I = G_m-G_l-G_r
\end{equation}

where \(G_m\) is the Gini impurity index of node \(m\), \(G_l\) and \(G_r\) are the Gini impurity index of the left and right split of node \(m\). Then, the Gini importance of feature \(X_j\) in a random forest model is calculated as Eq. \ref{eq:imp_rf},

\begin{equation}\label{eq:imp_rf}
I_j^{Gini} = \sum_{i=1}^{N}\sum_{m \in M_i} I_{ijm}
\end{equation}

where \(N\) is the number of trees, \(M_i\) is the set of nodes using feature \(X_j\) in tree \(i\). Last, normalized Gini importance is calculated as Eq. \ref{eq:nor_imp_rf},

\begin{equation}\label{eq:nor_imp_rf}
I_j = \frac{I_j^{Gini}}{\sum_i^F I_i^{Gini}}
\end{equation}

where \(F\) is the number of features.

The features are sorted by their Gini importance. The top seven features, about 20\% of the total, are deemed important for the model. Since the models are trained for specific county and environmental hazards, we add up the rankings of the same feature across different counties to establish an overall importance score \(I^{Overall}\). Specifically, \(I^{Overall}_j\), the overall importance score of feature \(X_j\) is calculated as Eq. \ref{eq:overall_imp}. 

\begin{equation}\label{eq:overall_imp}
I^{Overall}_j = \frac{\sum_{i=1}^C R_{i,j}}{F*C}
\end{equation}

where \(F\) is the number of features, \(C\) is the number of counties, \(R_{i,j}\) is the ranking of feature \(X_j\) in the i-th county based on normalized Gini importance. For example, \(R_{i,j}\) equals F if the normalized Gini importance of feature \(X_j\) in the i-th county is the highest among all features; on the other hand, \(R_{i,j}\) equals 1 if feature \(X_j\) has the least normalized Gini importance. The cumulative ranking is normalized by dividing it by the product of the number of features \(F\) and the number of counties \(C\), ensuring that the overall importance score \(I^{Overall}_j\) equals 1 if feature \(X_j\) has the highest normalized Gini importance in all counties.

From this aggregated score, we then select the most influential for each environmental hazard, i.e., for each hazard, the top seven features with highest overall importance score \(I^{Overall}\) will be assigned as the overall important features.

\section{Results}
\subsection{Model Predictability}
In this study, the predictability performance of models in determining the environmental hazard levels is used as the measurement of the extent to which environmental hazard disparity is shaped by variations of urban features. The better the predictability performance of models, the greater the role of urban feature variations in shaping environmental hazard exposure disparities. The rationale is that if the extent of hazards could be predicted accurately based on urban features related to the built environment, land cover, and socio-demographics, then variations of urban features could play an important role in shaping disparities in environmental hazard levels. Such an approach for measuring inequality has been used and validated; It is based on the premise that if hazard levels are not shaped by variations of urban features, a model would not be able to predict the hazard level based on features related to the built environment, human mobility, land cover, and socio-demographics c of areas. Accordingly, the predictability of model output indicates the extent to which variations urban features shape disparities in environmental hazard exposures of urban areas \cite{fan_inequality_2023}.

Similar to \cite{fan_inequality_2023}, this study uses F-score, the harmonic mean of
precision and recall, as the metric for model predictability:

\begin{equation}\label{eq:f-score}
F-score = (1+\beta^2) \cdot \frac{precision \cdot recall}{\beta^2 \cdot precision + recall}
\end{equation}

in which \(\beta\) can be selected such that recall is of \(\beta\) times important than precision. In this study, accurately identifying high-risk areas is of particular importance. Consequently, we use an F-score with \(\beta\) of 1.5 as the major indicator for evaluation of model performance to assign higher weights to recall over precision.

\begin{figure}[ht]
\centering
\includegraphics[width=0.8\textwidth]{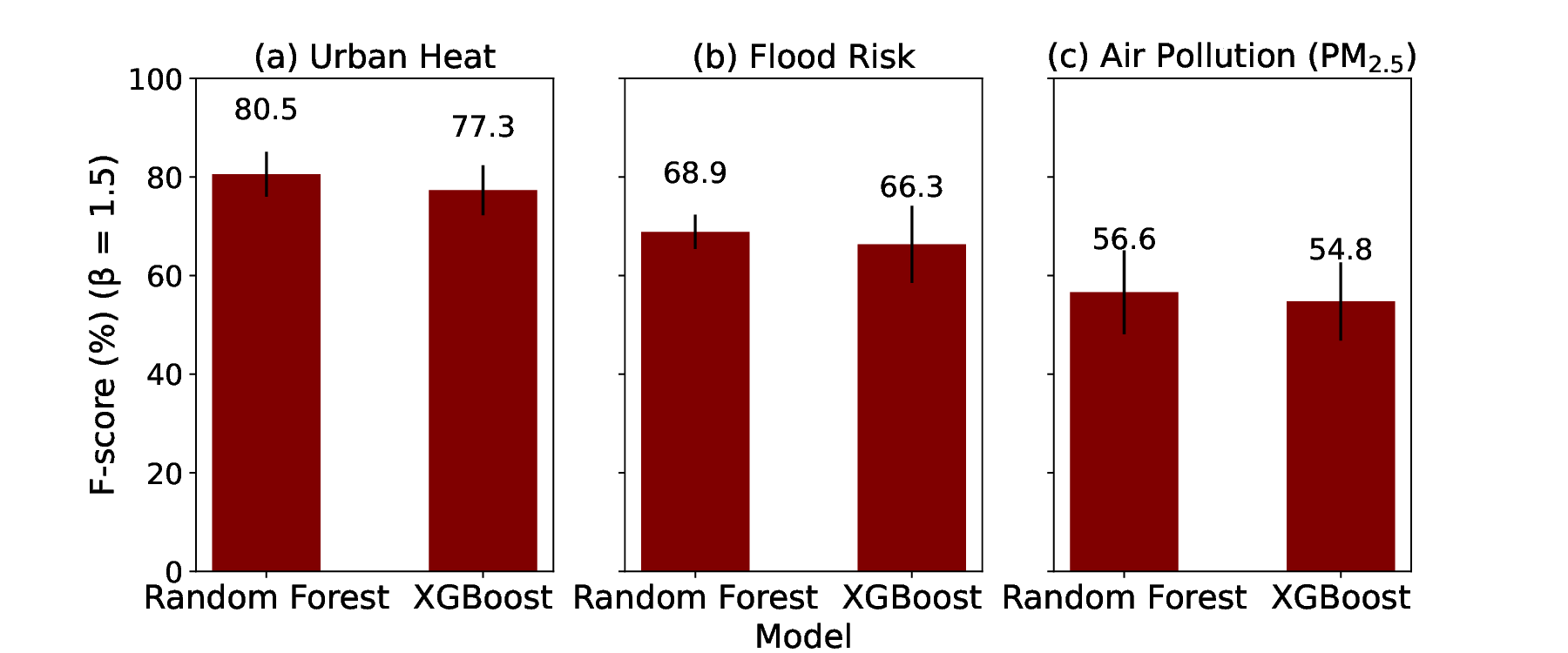}
\caption{Performance of random forest classifier and XGBoost classifier. (a) Average F-score of the random forest classifier and that of the XGBoost classifier for urban heat risk level prediction. (b) Average F-score for flood risk level prediction. (c) Average F-score for air pollution (PM\(_{2.5}\) concentration) prediction. (a)-(b) Average F-scores among six counties, whereas (c) Average F-scores among four counties due to data unavailability in Fulton County and Queens County. The \(\beta\) for the F-score is set as 1.5 for evaluation of all models. F-score is calculated based on test set. Both of the models yield the highest F-score for urban heat, followed by flood. In addition, random forest yields better prediction performances than XGBoost models for all three environmental hazards.}
\label{fig:performance}
\end{figure}

Random forest and XGBoost classifiers were trained and tested for the six counties. The average performance of random forest and that of XGBoost among the six counties is shown in Figure \ref{fig:performance}. The random forest yields better prediction performances than XGBoost models for all three environmental hazards. Consequently, the performance by random forest is used for subsequent analysis.

The model predictability among different regions is also compared. The performance of random forest models for the environmental hazards are presented in Table \ref{Tab:RF_perf}. Across all counties examined, the models demonstrate the highest predictability (F-score) for urban heat exposure. This suggests that the role of urban features in shaping the extent of urban heat is more pronounced compared to other hazards. On the other hand, the county achieving the highest predictability varies depending on the environmental hazard in question. For urban heat, Fulton County exhibits the highest predictability, implying a particularly strong interplay between urban feature variations and heat exposure in that region. Conversely, for flood risk and air pollution, the highest predictability is seen in Wayne County, and Harris County, respectively. These comparisons indicate that the influence of urban feature variations on each hazard is highly dependent on the contexts of specific county. These results highlight the imperative for deploying region-specific models and tailored intervention strategies to address the unique environmental challenges faced by each area.

\begin{figure}[ht]
\centering
\includegraphics[width=0.4\textwidth]{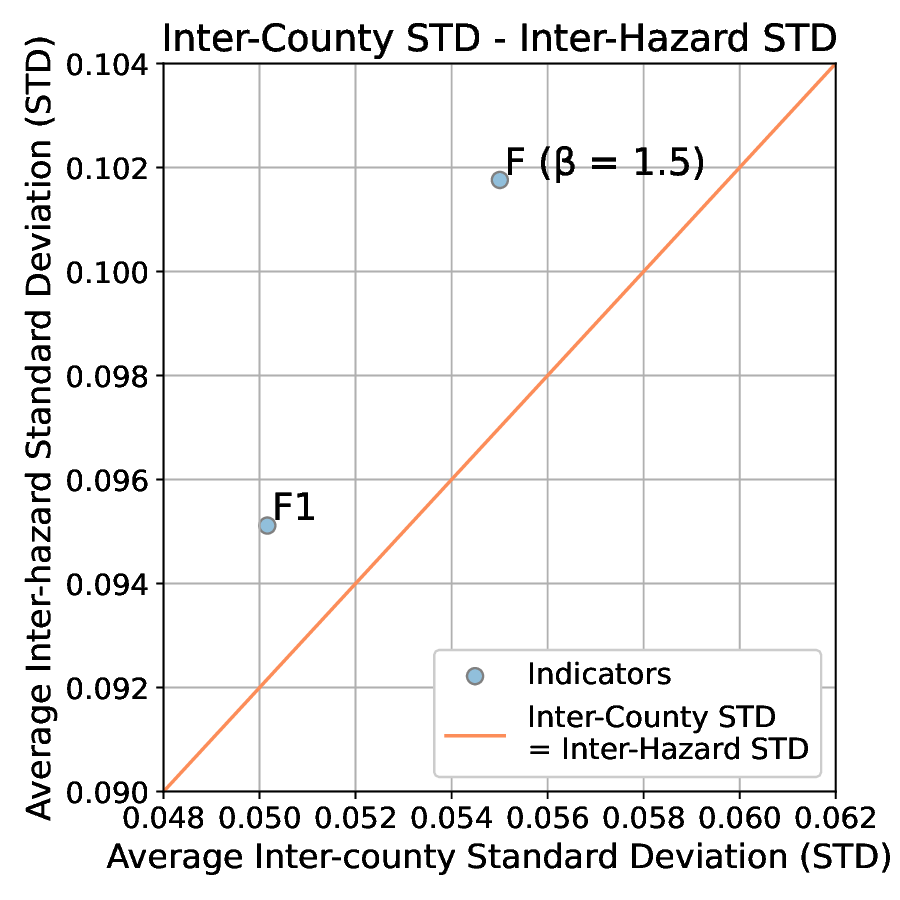}
\caption{Inter-county and inter-hazard standard deviation (\(\sigma^c\)/\(\sigma^h\)) distribution of the F-score representing the extent of the performance difference across different counties/hazards for the same hazard/county. The average inter-county standard deviation ($\overline{\sigma^c}$) is the average of values of the inter-county standard deviation (\(\sigma^c\)) of all hazards. .The average inter-hazard standard deviation ($\overline{\sigma^h}$) is the average of values of the inter-hazard standard deviation (\(\sigma^h\)) of all counties.}
\label{fig:city_hazard_std}
\end{figure}

We also compared inter-county and inter-hazard standard deviations of model performance to examine the change of the performance across different regions and hazards. The inter-county standard deviation captures the extent of performance differences for the same hazard across different counties. Similarly, the inter-hazard standard deviation captures the extent of the performance difference of hazards within the same county. The inter-county standard deviation \(\sigma^c\) of the performance for hazard \(h_j\) is calculated as Eq. \ref{eq:std_county}. 

\begin{equation}\label{eq:std_county}
\sigma^c_j = \sqrt{\frac{\sum_{i=1}^C (M_{i,j}-\mu^c_j)^2}{C}}
\end{equation}

where \(C\) is the number of counties, \(M_{i,j}\) is the value of a metric of the model for county \(c_i\) and hazard \(h_j\), \(\mu^c_j\) is the mean value of a metric of the models for hazard \(h_j\) across all counties. The inter-hazard standard deviation \(\sigma^h\) of the performance for county \(c_i\) is calculated as Eq. \ref{eq:std_hazard}. 

\begin{equation}\label{eq:std_hazard}
\sigma^h_i = \sqrt{\frac{\sum_{j=1}^H (M_{i,j}-\mu^h_i)^2}{H}}
\end{equation}

where \(H\) is the number of hazards, \(\mu^h_i\) is the mean value of a metric of the models for county \(c_i\) across all hazards. We then take an average of inter-county standard deviations \(\sigma^c\) across all hazards and the average of inter-hazard standard deviations \(\sigma^h\) across all counties. Figure \ref{fig:city_hazard_std} shows the distribution of the average inter-county and inter-hazard standard deviations of F-score. 

For F1, the average inter-county standard deviation $\overline{\sigma^c}$ and average inter-hazard standard deviation $\overline{\sigma^h}$ are respectively 0.050 and 0.095. For F (\(\beta\) = 1.5) , the average inter-county standard deviation $\overline{\sigma^c}$ and average inter-hazard standard deviation $\overline{\sigma^h}$ are respectively 0.055 and 0.102. This result shows the inter-hazard standard deviation of the F-score is about twice as great than that of inter-county standard deviation. The inter-county difference in model predictability is less significant than the inter-hazard difference, suggesting that the extent to which urban features shape spatial hazard exposure for the same hazard across different counties is similar. Conversely, the extent to which variations of urban features shape spatial hazard exposure for different hazards in a single county vary. For instance the extent to which urban features shape disparities in urban heat hazard levels is greater than for flood and air pollution hazards.

\begingroup

\renewcommand{\arraystretch}{1.5} % Default value: 1

\begin{table}
\setlength\tabcolsep{0.5pt}
\caption{Random Forest Model Performance (\%)}
\centering
\begin{tabular*}{\textwidth}{@{\extracolsep{\fill}}*{13}{c}}
\toprule
& \multicolumn{3}{c}{Training Data Size} & \multicolumn{3}{c}{Test Data Size} & \multicolumn{3}{c}{F1}& \multicolumn{3}{c}{F (\(\beta = \)1.5)} \\
\cline{2-4} \cline{5-7} \cline{8-10} \cline{11-13}
& Heat & Flood& Air*& Heat & Flood & Air & Heat& Flood & Air & Heat& Flood& Air\\
\midrule
Harris, TX& 670& 670& 110& 288& 288 & 48& 81.97 & 66.42 & 69.57 & 84.09 & 65.14& 68.42\\
Cook, IL& 782& 782& 233& 336& 336 & 101 & 75.61 & 70.17 & 54.79 & 74.91 & 72.16& 56.58\\
Wayne, MI & 319& 319& 54 & 138& 138 & 24& 80.68 & 72.22 & 48.00 & 82.85 & 73.40& 48.75\\
Suffolk, MA & 130& 130& 29 & 56 & 56& 13& 79.31 & 65.45 & 54.55 & 80.38 & 65.00& 52.70\\
Fulton, GA& 179& 179& \diagbox[width=2em]{}{}& 77 & 77& \diagbox[width=2em]{}{}& 83.33 & 67.50 & \diagbox[width=2em]{}{}& 85.69 & 68.82& \diagbox[width=2em]{}{}\\
Queens, NY& 373& 373& \diagbox[width=2em]{}{}& 160& 160 & \diagbox[width=2em]{}{}& 74.85 & 67.88 & \diagbox[width=2em]{}{}& 75.38 & 68.68& \diagbox[width=2em]{}{}\\
Average & 409& 409& 107& 176& 176 & 47& 79.29 & 68.27 & 56.73 & 80.55 & 68.87& 56.61\\ 
\bottomrule
\multicolumn{13}{l}{*Air in this table refers to air pollution.}\\
\end{tabular*}
\label{Tab:RF_perf}
\end{table}

\endgroup

\subsection{Feature Importance Analysis}
\begin{figure}[ht]
\centering
\includegraphics[width=1\textwidth]{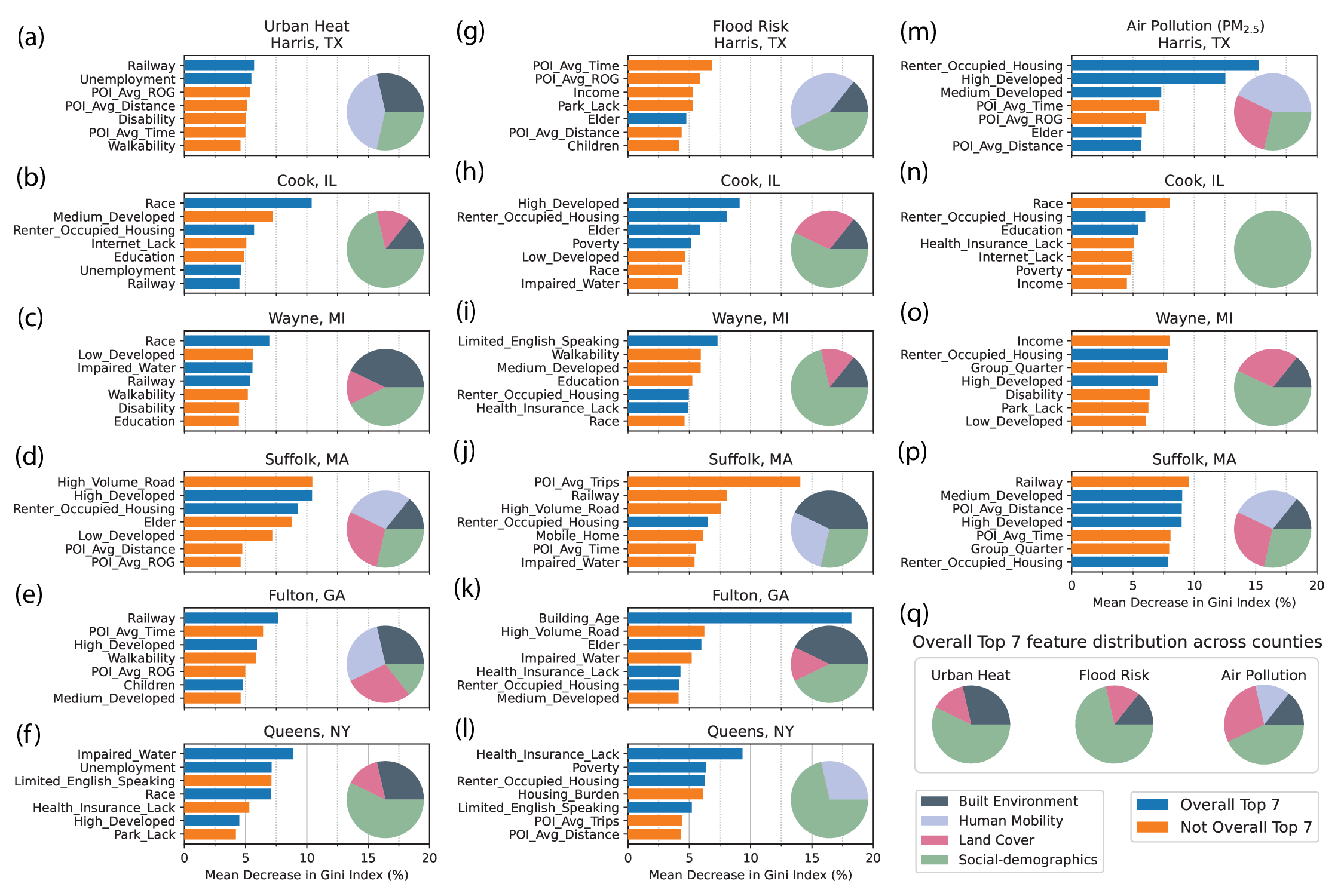}
\caption{Feature importance and the important feature distribution for each environmental hazard and each county. (a)-(f) Top seven important features for urban heat risk in each county and the proportions of different feature domains in top seven features. (g)-(l) Top seven important features for flood risk and the proportions of different feature domains in top seven features. (m)-(p) Top seven important features for air pollution (PM\(_{2.5}\) concentration) and the proportions of different feature domains in the top seven features. (q) Proportions of different feature domains in overall top seven important features to each environmental hazard across all counties. The listed features in (a)-(p) are the respective important features in each county; each feature is labeled as to whether it is among the top seven important features in the overall ranking across all counties.}
\label{fig:feature_importance}
\end{figure}

\begin{figure}[ht]
\centering
\includegraphics[width=1\textwidth]{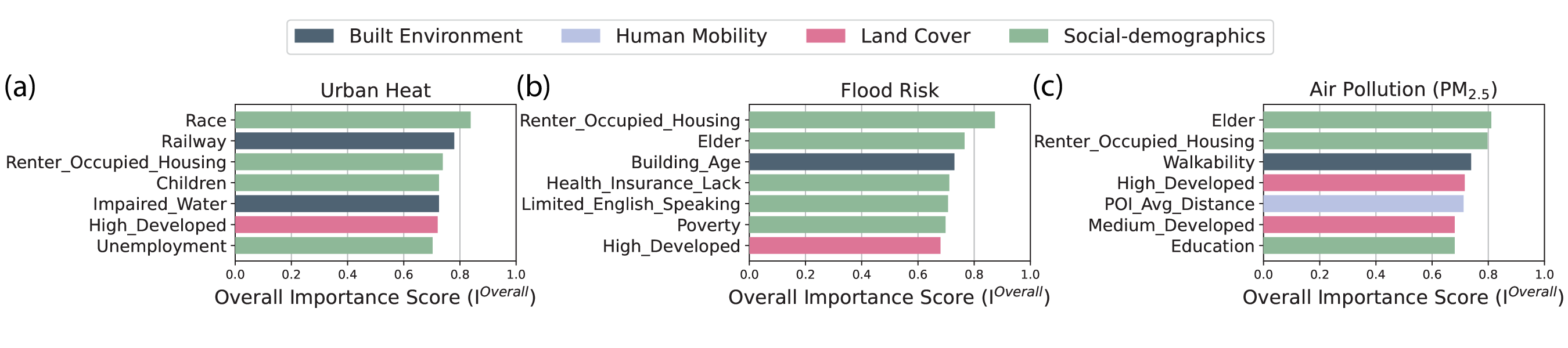}
\caption{Overall importance score of top seven overall important features across counties for each environmental hazard. (a) Top seven overall important features for urban heat risk. (b) Top seven overall important features for flood risk. (c) Top seven overall important features for air pollution (PM\(_{2.5}\) concentration). Overall importance score \(I^{Overall}\) is calculated as Eq. \ref{eq:overall_imp}.}
\label{fig:feature_importance_overall}
\end{figure}

The important urban features in predicting the environmental hazard levels in each county are presented in Figure \ref{fig:feature_importance}. The overall important features across counties for each environmental hazard are presented in Figure \ref{fig:feature_importance_overall}. For all three environmental hazards, social-demographic characteristics is the dominant important feature group in the top seven overall important features across counties. Specifically, the most important features for urban heat, floods, and air pollution are respectively the percentage of non-white population, the percentage of renter-occupied housing units, and the percentage of the elder population. These results reveal significant disparities in hazard exposures by vulnerable sub-populations.

Besides the significant importance of social-demographic features, the importance of other feature groups vary for different hazard types. Built environment features, especially transportation infrastructure features, are of greater importance for urban heat than for other hazards, whereas human mobility and land cover features are relatively more important for air pollution. In addition, although social-demographics is the most important feature group in general, it does not dominate the top 7 features in some counties. In Harris County, human mobility features are prevalent among the important features for urban heat and air pollution (PM\(_{2.5}\) concentration), while social-demographics and human mobility features share equivalent importance in predicting flood hazard exposure levels. In Wayne County, social-demographics and built environment features have equal importance in predicting urban heat levels. In contrast, built environment features are among the important features contributing to flood hazard exposure in Suffolk County. While for urban heat and air pollution, features from various domains—including the built environment, human mobility, land cover, and socio-demographics—exhibit similar levels of importance in shaping exposure to these hazards. Likewise, in Fulton County, each feature group also carries a similar weight for predicting urban heat levels. The presence of built environment and land cover features among the important features in predicting hazard levels provides important evidence for the role of urban features in shaping disparities in environmental hazard levels across different areas. 

The results also reveal that the contributions of urban features vary for each hazard types and across counties. Social demographics consistently emerge as the most influential feature group across various counties and hazards. This observation underscores the widespread issue of environmental injustice affecting socially vulnerable communities. However, certain counties also exhibit unique sets of important features, suggesting that localized forms of inequality are at play alongside the more universal factors. This multifaceted landscape points to the need for targeted urban development strategies based on the specific urban features of individual counties. For example, in Harris County, human mobility features are identified as the primary feature group affecting all three hazards. This pattern may be attributable to the decentralized urban structure prevalent in Texas, and given Harris County's extensive geographical scope, the role of human mobility features becomes notably accentuated. The significance of human mobility and social demographics in these models underscores the need for targeted urban development strategies to alleviate disparities in hazard exposures. Creating more inclusive and equitable urban structures, such as better public transportation options and improved accessibility to amenities, could serve as vital steps in alleviating urban features that shape disparities in environmental hazards exposure. In Suffolk County, the importance of the four feature groups is more evenly distributed compared to other counties. This implies that hazard exposure disparities in Suffolk County are shaped by a broader range of heterogeneous urban features. The balanced feature importance highlights the need for a integrated urban design strategies strategy for alleviating environmental injustice issues. Addressing various aspects of urban design and planning, from enhancing public transportation to refining land-use policies to improving access to facilities and equitable infrastructure, can offer a more holistic solution to alleviating environmental injustice in this county. By focusing on these key and localized features, urban planners and policymakers can more effectively address both universal and locality-specific determinants of environmental hazard exposure disparities.

\subsection{Model Transferability}
\begin{figure}[ht]
\centering
\includegraphics[width=1\textwidth]{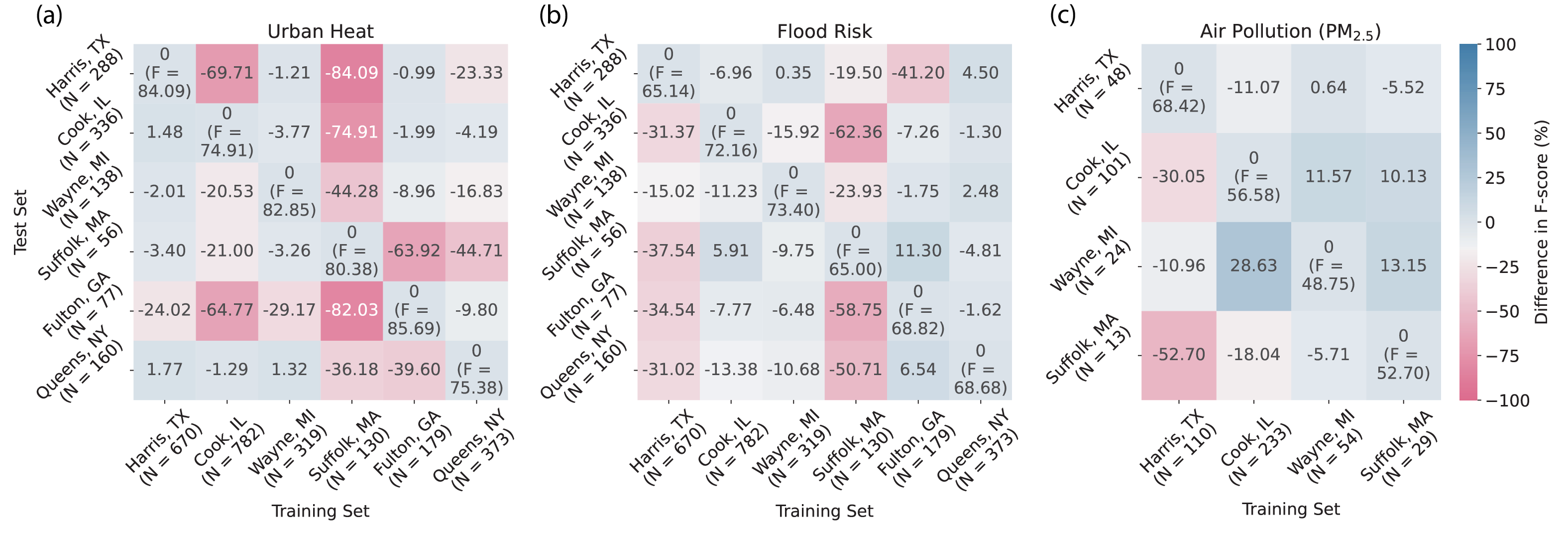}
\caption{Cross-county performance difference. (a) Difference of F-score when using the urban heat risk model trained in one county to test in another county. (b) Difference of F-score when using the flood risk model trained in one county to test in another county. (c) Difference of F-score when using the air pollution (PM\(_{2.5}\) concentration) model trained in one county to test in another county. \(\beta\) for F-score is 1.5. Blue shading represents transferable, whereas red color represents untransferable.}
\label{fig:crosscity}
\end{figure}

\begin{figure}[ht]
\centering
\includegraphics[width=1\textwidth]{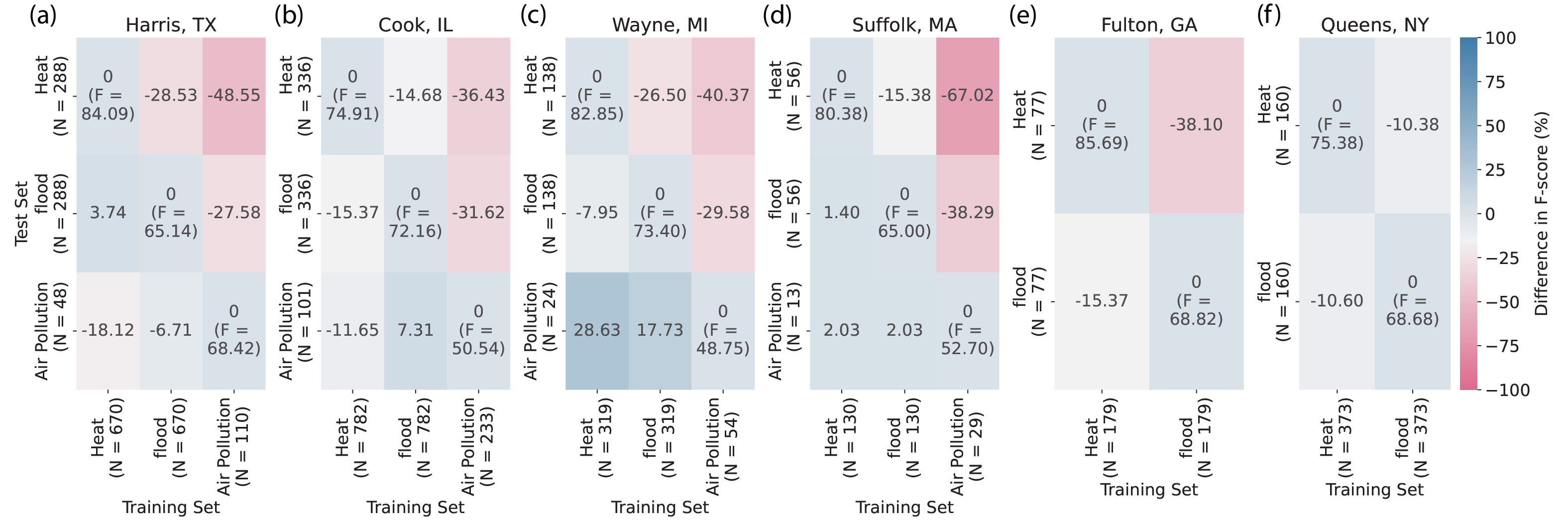}
\caption{Cross-hazard performance difference. (a) Difference in F-score when using the model trained for one environmental hazard to predict another hazard, in Harris County. (b) Difference in F-score in Cook County. (c) Difference in F-score in Wayne County. (d) Difference in F-score in Suffolk County. (e) Difference in F-score in Fulton County. (f) Difference in F-score in Queens County. \(\beta\) for F-score is 1.5. Blue shading represents transferable, whereas red color represents not transferable.}
\label{fig:crosshazard}
\end{figure}

Model transferability is a function of the generalizability of the model and effectiveness of applying findings from one region to other regions. In this step, we examine the model transferability by evaluating model's cross-county and cross-hazard performance (Figure \ref{fig:crosscity} and Figure \ref{fig:crosshazard}). These figures show the performance difference of the model when trained in one county/hazard and tested in another county/hazard. For instance, the first column in Figure \ref{fig:crosscity} (a) illustrates the difference in F-score when utilizing the urban heat model trained in Harris County, both for testing within Harris County and in other counties. Similarly, the first column in Figure \ref{fig:crosshazard}(a) denotes the discrepancy in F-score when the model trained for urban heat in Harris County is employed to test urban heat data within the same county versus when it is used to test data for other hazards within Harris County. We establish the transferability threshold for the F-score difference at -15\% . Thus, a model is deemed transferable to another region or hazard type if the test on that dataset yields an F-score decrease of no more than 15\% compared to the result derived from the model both trained and tested on that same dataset. The rationale of setting the threshold at -15\% level is that a decrease of 15\% still better than the expected F-score of a random binary classifier, since the F-score of our experiments mostly lie in 65\% to 80\%.

As shown in Figure \ref{fig:crosscity} (a) and Figure \ref{fig:crosscity} (b), the models trained in Suffolk County for urban heat and flood risk and the model trained in Harris County for flood risk fail when transferred to all other counties. These results are consistent with the findings of feature importance results presented in Figure \ref{fig:feature_importance}. Most of the important features in Suffolk County for urban heat and flood risk and those in Harris County for flood risk do not match the overall important features (see Section 2.3). Even the important features for the model of predicting Harris county air pollution has a large overlap with the overall important features, it still fails to be transferred to most of the counties. This could be due to the possibility that the feature importance pattern for air pollution (PM\(_{2.5}\)) in Harris County differs significantly from patterns observed in other counties. In Harris County, the top two features hold exceptionally high importance levels of more than 15\% and about 12.5\%, respectively. In contrast, all other features register less than 10\% importance. Meanwhile, in other counties, the importance levels of the top seven features are fairly balanced, each falling below 10\%. This pronounced disparity in feature importance makes the air pollution model developed for Harris County less generalizable to other areas

Regarding, cross-hazard model transferability, Figure \ref{fig:crosshazard} shows that the models for flood risk and air pollution fail to be transferred to urban heat in most counties, while the models for urban heat are able to be transferred for predicting other hazards in most counties. In some cases, the models trained for urban heat even achieve slightly better performance in predicting flood risk and air pollution than the models trained for flood risk and air pollution. The models for flood risk can be transferred for predicting air pollution, yet are non-transferable for urban heat in most counties. This result implies that urban features that shape the extent of urban heat hazard level may also shape hazard levels for flood and air pollution. Hence, strategies targeting features to alleviate urban heat hazards could also alleviate flood and air pollution hazards. Accordingly, by alleviating urban heat hazard, cities can harness co-benefits and reduce flood and air pollution hazard levels as well.

\section{Discussion}

\subsection{Predictability Analysis of Hazard Exposures and Inequality}
Our results demonstrate that both random forest and XGBoost can predict urban heat, flood risk, and air pollution with satisfactory accuracy, using the various urban heterogeneity as inputs. The results demonstrate that the urban features not only shape the city structures, but also significantly influence the emergence, progression, and mitigation of environmental hazard exposures, subsequently causing the inequality of hazard exposures within the study regions. Although higher predictability performance suggests a greater extent of hazard exposure disparity due to urban features, lower performance for air pollution prediction compared with flood risk and air pollution hazards may also be due to the limited training dataset for PM\(_{2.5}\) concentration. Moreover, both methods achieve the highest accuracy for the prediction of urban heat exposure (F-score about 80\%). Such high predictability indicates that the extent of disparity in urban heat exposure is particularly shaped by urban features. The inter-county and inter-hazard performance comparisons further indicate that the role of urban features in shaping environmental hazard exposure disparities varies across hazards but is notably consistent within the same type of hazard across different counties. Such consistency in inequality within the same type of hazard across different counties suggests that there may be overarching, systemic determinants influencing these specific types of environmental hazard exposures, irrespective of local contexts. On the other hand, the variation in the extent of exposure disparities across different hazards suggests that each type of hazard might be influenced by a distinct set of urban features, requiring targeted policies for alleviating exposures for each hazard type. 

\subsection{Causal Relationships Interpretations Between Urban Features and Environmental Hazards}
The relationships between urban heterogeneous features and environmental hazards exposure may result from three potential cases: (1) both the urban features and hazard exposures originate from a shared root cause, (2) the exposure to hazards is a direct outcome of particular urban features, or (3) the urban features develop as a result of hazard exposure. The finding that race distribution is the most important features for urban heat exposure may probably belong to the first case, which mean that certain racial groups are more inclined to work in urban or industrial areas and consequently give rise to race distribution's important contribution to urban heat. Similarly, the higher importance of transportation infrastructure on urban heat may also be due to the possibility that the development of transportation infrastructure and urban heat having the same cause—urbanization. In contrast, the higher importance of human mobility on air pollution shows the influence of traffic on air pollution, which may follow the second case, i.e., hazard exposure resulting from urban features. On the other hand, the percentage of renter-occupied housing units as the most important feature of flood risk may be the third case, i.e., urban features resulting from hazard exposure. High flood risk can result in low property value, thereby prompting landlords to buy these houses for rent.
\subsection{Assessing Model Transferability Across Counties for Informed Urban Development Strategies}
The results for model transferability across counties align well with the feature importance analysis. Given that both universal and localized exposure disparity issues are at play, models developed in one county often do not translate directly to others. For example, urban heat and flood risk models for Cook County and Queens County are two-way transferable, meaning they can be applied interchangeably. Conversely, the urban heat models for Fulton County are transferable to other counties, but not the other way around. In other instances, such as with the urban heat and flood risk models for Suffolk County, there is no transferability at all. This variability in model transferability has important implications for the universality of urban development strategies for alleviating exposure disparities. When models are two-way transferable, shared interventions across similar urban environments could be more readily implemented. On the other hand, non-transferable models indicate the need for highly localized strategies tailored to each county's unique set of challenges. This fluctuation in model transferability has direct consequences for urban development policy. For counties where models are two-way transferable, such as Cook and Queens counties, effective strategies and solutions for tackling similar hazards could be shared. This might include initiatives such as greening urban spaces to mitigate both heat and flood risks simultaneously or implementing smart traffic systems to reduce human mobility-related hazards. In contrast, non-transferable models, such as those for Suffolk County, call for highly localized interventions. Such strategic planning should also be adaptive and flexible to account for the unique challenges each county faces, whether it be demographic diversity, specific topographical features, or differing levels of industrial activity. The non-transferability of models emphasizes the importance of community-specific data and local context in shaping effective, equitable urban development strategies that could alleviate environmental hazard exposure disparities. Model transferability across different hazards also correlates with the variance in model performance among them. Utilizing models trained with high-quality data from one hazard can sometimes yield good performance when applied to other hazards with limited data. This underlines the critical role of data quality in achieving robust model training, suggesting that investing in high-quality data collection could benefit multiple areas of hazard management and urban planning

\section{Closing Remarks}
\label{sec:5}
The inequality in exposure to environmental hazards has long been a concerning urban issue, potentially further aggravating the living conditions of vulnerable communities. This paper provides a novel analysis for urban features' role in influencing the disparities in exposures to various environmental hazard using interpretable machine learning. Our analysis and results affirm the significance of urban features in shaping environmental hazards and its corresponding injustice issue. The results confirm the presence of dire disparities with race, renter-occupied housing, and the elder population as the most important features, respectively, for predicting urban heat, floods, and air pollution. The varying degrees of model transferability across counties serve to inform the complexity and diversity of these local realities, underscoring the need for data-driven, localized urban development strategies and policies for addressing environmental hazard exposure inequality and equitable urban design strategies.

From a theoretical perspective, this research brings fresh insights to the discourse on environmental justice research by placing a keen focus on the role and impact of urban features and their complex interactions. This work straddles the intersection of environmental justice theories, urban science, and machine learning to deliver a deeper analysis of the urban factors shaping environmental hazard exposure disparities. The interpretable machine learning models illuminate the complex dynamics at play, giving a fresh perspective on the influence exerted by urban characteristics on hazard exposure disparities. This study adds a valuable layer of understanding to existing theories, acknowledging that economic, sociopolitical, and racial factors are entwined with the built environment and human mobility in producing environmental hazard exposure inequality. 

On a practical plane, our research carries weighty implications for environmental policy and urban planning. By revealing how various urban features contribute to disproportionate exposure to environmental hazards, the study can provide urban planners with a data-driven and analytics-based tool to understand the role of various urban development strategies on the environmental hazards exposures and to guide policy interventions to rectify associated disparities. With the detailed insights obtained from the machine learning models, policymakers can devise integrated urban design strategies that take into account the complex socio-spatial dynamics at play in environmental injustice. Moreover, the investigation into model transferability can support the creation of broader, adaptable environmental policies that can be fine-tuned to different regional contexts. Therefore, this research not only helps unravel the relationship between urban features and environmental hazard exposure disparities but also sets the groundwork for the development of data-driven and analytics-informed solutions.

Despite the significance of our findings, our study has certain limitations that offer potential avenues for future research. First, the selection of six counties, albeit diverse, may not fully capture the complexity of environmental injustice in different geographical and cultural contexts, underscoring the need for a broader scope in future studies. Second, while our machine learning models are interpretable, they may not fully unravel the multifaceted nature of environmental injustice, which may be shaped by intricate and often intangible social processes beyond the scope of measurable urban features. Third, while we explored a variety of urban features, other potential factors, such as policy decisions and historical contexts, were not fully considered. Finally, the transferability of our findings may be affected by unique regional factors that could cause variations in environmental injustice extent. Future research could build upon our models and the findings and examine a wider range of determinants and also test the models in other regions for a more nuanced understanding of the complex urban dynamics that shape environmental injustice in cities and communities.

\section{Data Availability}
The data that support the findings of this study available from SafeGraph, Spectus, and First Street Foundation were used under license. These data can be accessed upon request submitted to the data providers. Other data we use in this study are all publicly available.

\section{Code Availability}
The code that supports the findings of this study is available from the corresponding author upon request.

\section{Acknowledgements}

\bibliography{ref}

\end{document}